\documentclass[11pt]{article}

\usepackage[preprint]{acl}
\usepackage{enumitem}
\usepackage{algorithm}
\usepackage{algorithmic}
\usepackage{times}
\usepackage{latexsym}
\usepackage{booktabs}
\usepackage[T1]{fontenc}

\usepackage[utf8]{inputenc}

\usepackage{microtype}

\usepackage{inconsolata}
\usepackage{subcaption}

\usepackage{amsmath}
\usepackage{graphicx}
\usepackage{tcolorbox}

\usepackage{inconsolata}
\usepackage{xcolor}
\usepackage{xspace}
\usepackage{hyperref}
\usepackage{cleveref}

\newcommand{\sys}[0]{\texttt{ACONIC}\xspace}
\newcommand{\sysfull}[0]{\texttt{Analysis of CONstraint-Induced Complexity}\xspace}

\definecolor{grey}{rgb}{0.5,0.5,0.5}

\title{An Approach for Systematic Decomposition of Complex LLM Tasks}

\author{
  Tianle Zhou$^{\S}$,
  Jiakai Xu$^{\S}$,
  Guanhong Liu$^{\S}$,
  Jiaxiang Liu$^{\S}$,
  Haonan Wang$^{\S}$,
  Eugene Wu$^{*{\S}}$ \\
  $^{\S}$Columbia University \\
}

\begin{document}
\maketitle
\begin{abstract}
Large Language Models (LLMs) suffer from reliability issues on complex tasks, as existing decomposition methods are heuristic and rely on agent or manual decomposition. This work introduces a novel, systematic decomposition framework that we call \sysfull (\sys), which models the task as a constraint problem and leverages formal complexity measures to guide decomposition.   On combinatorial (SATBench) and LLM database querying tasks (Spider), we find that by decomposing the tasks following the measure of complexity, agent can perform considerably better.
\end{abstract}

\section{Introduction}

Large Language Models (LLMs) have demonstrated impressive competence across a wide range of reasoning, programming, and problem-solving tasks. Yet, when faced with \emph{complex tasks} that require deep multi-step reasoning or combinatorial search, even state-of-the-art models often fail to produce correct results in a single forward pass. 

\begin{figure}
    \centering
    \includegraphics[width=\linewidth]{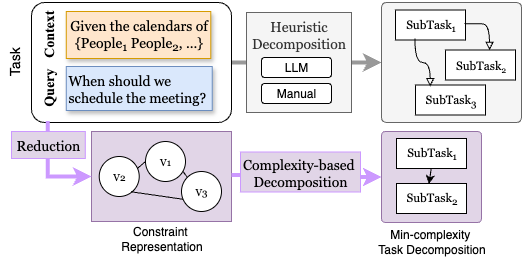}
    \caption{Given a scheduling task, existing approaches use heuristic LLM or manual task decomposition methods (\textcolor{gray}{top row}).  Our framework reduces the task into a constraint satisfaction problem that allows for a systematic decomposition that minimizes the problem complexity (\textcolor{purple}{bottom row}).      }
    \label{fig:overview}
\end{figure}

A growing body of work addresses this limitation through \emph{task decomposition}. Instead of solving a task monolithically, these methods break it into smaller, more tractable subtasks.  One popular line of work, starting from chain-of-thought~\cite{wei2022chain}, is to use LLMs for decomposition~\cite{yao2023tree,khot2023decomposed, pourreza2023dinsql, chen2024divide}.  Other methods rely on domain experts to decompose the task into workflows and provide access to tools that shoulder parts of the task~\citep{wang2024executable, singh2024agentci}.  We summarize these methods as the \textcolor{gray}{grey path} in \Cref{fig:overview}.

While decomposition seeks to break more complex tasks into workflows of simpler subtasks, existing approaches are largely heuristic.  When is a task ``complex''?   How should it be decomposed?   
A principled measure of task complexity would enable systematic decomposition strategies and the ability to study tasks of comparable difficulty, and provide guidance on when tools are needed.


In this paper, we introduce a formal complexity framework for LLM tasks by reducing them into \emph{constraint satisfaction problems} (specifically, 3-SAT). We use properties of the induced constraint graph (graph size and treewidth) as measures of task complexity. Building on these measures, we propose a decomposition method based on ~\cite{bodlaender1998partial} that minimizes subtask complexity under this formalization, yielding decompositions that preserve global satisfiability while maximizing local solvability.  Along the \textcolor{purple}{purple path} in \Cref{fig:overview}, we first reduce the task---modeled as a context that describes a set of constraints and a query that must reason over the constraints---into a formal constraint satisfiability problem.   We then decompose the constraint problem and construct a workflow over the subtasks defined for each subproblem.

We use \textsc{SAT-Bench}~\citep{wei2025satbenchbenchmarkingllmslogical} and  the NL2SQL \textsc{Spider}~\citep{yu2019spiderlargescalehumanlabeleddataset} benchmark to study this framework.  We find that the task measures define \textbf{frontiers of difficulty} that separate tasks easily solvable by LLMs from those that are nearly impossible without structural assistance. In addition, complexity-guided decomposition improves completion on SATBench and outperforms Tree-of-Thoughts on Spider by 3–8 accuracy points across difficulty levels.

\section{Method}



At a high level, we find that a class of agent tasks that admit constraint formulations can be reduced to constraint satisfaction problems (CSPs), such that solving the CSP is equivalent to solving the original task. This formulation makes explicit the dependency structure among task variables, which allows us to leverage tree decomposition from constraint processing theory to decompose complex tasks into smaller, weakly coupled subproblems that can be reasoned about locally and composed consistently.

\subsection{Reduction to Constraint Satisfaction}

We model a class of agent tasks as instances of constraint satisfaction problems (CSPs). A CSP is defined by a tuple $\mathcal{C} = \langle X, D, \mathcal{R} \rangle$, where $X$ is a finite set of variables, $D$ assigns each variable a finite domain, and $\mathcal{R}$ is a set of constraints over subsets of variables.

Within this framework, the \emph{state} of the agent is a (partial) assignment to variables in $X$, and an \emph{action} corresponds to proposing or updating values for a subset of variables, or acquiring information that refines the constraints. The agent’s objective is to construct a complete assignment that satisfies all constraints in $\mathcal{R}$.

We illustrate this abstraction using a simple meeting scheduling example.
\begin{tcolorbox}[colback=black!10!white, colframe=white, boxrule=0pt]\small
\emph{Motivation Example — Meeting Scheduling.}

Alice needs to meet with Bob and Charlie separately and asks an agent to schedule both meetings. 
The system records show:
\begin{itemize}[leftmargin=*,itemsep=0em]
  \item Alice: \texttt{["Alice is available at morning in office r", "Alice is available in the afternoon at office t"]}
  \item Bob: \texttt{["Bob is available in the afternoon at office t"]}
  \item Charlie: \texttt{["Charlie is available at morning in office r", "Charlie is available in the afternoon at office t"]}
\end{itemize}
\textbf{Task.} How should the agent compose the invitation emails to Alice--Bob and Alice--Charlie?

\end{tcolorbox} 
In this example, resolving conflicts is non-trivial because assignments for one meeting constrain the feasible choices for the other. For instance, choosing a time for Alice and Charlie may invalidate the only feasible option for Alice and Bob.

We introduce Boolean variables $x_{p}^{(t,\ell)}$ indicating whether person $p$ attends location $\ell$ at time $t$. Let $\mathcal{P}$ denote the set of participants, $\mathcal{T}$ time slots, and $\mathcal{L}$ locations. For each person $p$, let $\mathcal{A}_p \subseteq \mathcal{T} \times \mathcal{L}$ denote feasible slots, and for each pair $(p,p')$ let $\mathcal{F}_{p,p'}=\mathcal{A}_p \cap \mathcal{A}_{p'}$.

The goal is to select exactly one feasible meeting slot for each desired pair $\mathcal{F}_{A,B}$ and $\mathcal{F}_{A,C}$, and to ensure that no one attend two meetings simultaneously. This leads to a set of constraints over the variables $\{x_{p}^{(t,\ell)}\}$ (As demonstrated in appendix \ref{app:meeting_sat_to_csp}).

The agent must produce a valid initiation of the variables $X$ that satisfies all the constraints.
Our objective is to find a problem decomposition strategy that provide the agent with observations over a subset of the problem.

\subsection{Tree Decomposition for Agent Tasks}
\begin{figure}
    \centering
    \includegraphics[width=\linewidth]{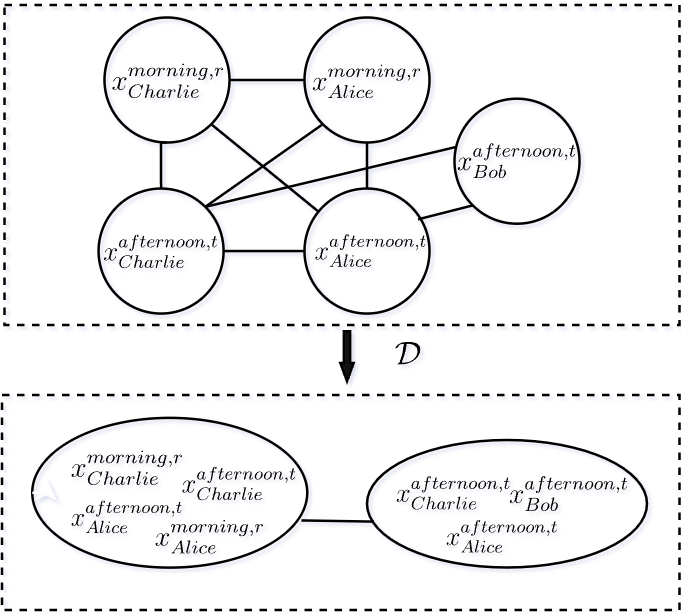}
    \caption{CSP graph (and its tree decomposition) for Meeting Scheduling example. Constructed based on CSP representation in appendix \ref{app:meeting_sat_to_csp}}
    \label{fig:decomp_sample}
\end{figure}
As illustrated in \ref{fig:decomp_sample}, the CSP induces a constraint graph $G = (X,E)$, where variables are nodes and an edge connects two variables if they co-occur in some constraint. It provides a structural abstraction of the task, capturing which task variables are mutually dependent through constraints. This representation allows us to reason about task structure independently of concrete values, and motivates decomposition strategies that group tightly coupled variables into localized subproblems. We therefore explore how tree decomposition~\cite{bodlaender1998partial} of this graph can be used to organize and simplify agent reasoning.

Tree decomposition maps $G$ into a tree of bags $\mathcal{D} = (T, \{B_i\})$, where each bag $B_i \subseteq X$ satisfies:
\begin{itemize}
  \item Every constraint is contained in some bag.
  \item For any variable $x$, the set of bags containing $x$ forms a connected subtree.
\end{itemize}

Following these properties, tree decomposition ensures that if each bag enforces local consistency on its induced constraints, then these local solutions can be merged into a global satisfying assignment as long as shared variables agree along the tree.

Among all tree decompositions $\mathcal{D}=(T,\{B_i\})$ of $G$, there exists an optimal decomposition
\(
\mathcal{D}^* \in \arg\min_{\mathcal{D}} \ \max_{i} |B_i|.
\)
The corresponding minimum possible maximum bag size,
\(b^*(G) := \min_{\mathcal{D}} \max_i |B_i|,\)
induces the \emph{treewidth} of $G$ as $\mathrm{tw}(G)= b^*(G)-1$. We use this quantity to characterize the intrinsic structural complexity of the underlying constraint problem: smaller $\mathrm{tw}(G)$ implies that the task can be decomposed into smaller, more weakly-coupled subproblems.

We exploit this structure to decompose reasoning into minimal, locally consistent subtasks. Each bag corresponds to a subproblem over a small variable set. The agent observes and reasons following the subproblems, while keeping consistency over intersections and ensuring global satisfiability. This leads to a principled decomposition strategy whose complexity is governed by the intrinsic treewidth of the task.

\section{Experiments}
We now evaluate our framework against  chain-of-thought decomposition~\cite{wei2022chain} and Tree of Thoughts~\cite{yao2023tree} on two datasets: SAT-based story problems (SAT-bench) and natural language to SQL (Spider).   


\subsection{SAT-Bench}
SAT-Bench~\cite{wei2025satbenchbenchmarkingllmslogical} uses  
(i) an underlying SAT problem to construct
(ii) a natural-language story describing the same constraints, and 
(iii) an alignment from story entities to their SAT representations. 
Instead of asking the LLM for an immediate satisfiability judgment, 
we prompt it to produce value assignments step-by-step to ensure a complete understanding of the assignment process. 
In this process, the agent’s state is a partial assignment to SAT variables, and progress corresponds to extending this assignment while maintaining consistency with the underlying constraints. Concretely, we treat the SAT-Bench CNF as a Boolean CSP, where the goal is to find an assignment that satisfies all clauses.

At each round of variable assignment, the agent is provided with the story, the variable to SAT mapping rules, and a set of observations. We tested agent with two different configurations of observations. 
In the chain-of-thought baseline,
the agent observes all condition observations.
In the tree-decomposition setup, the agent instead observes all conditions belonging to the same 
subproblem induced by the decomposition.


\begin{figure}[t]

    \begin{subfigure}[t]{0.49\linewidth}
        \centering
        \includegraphics[width=\linewidth]{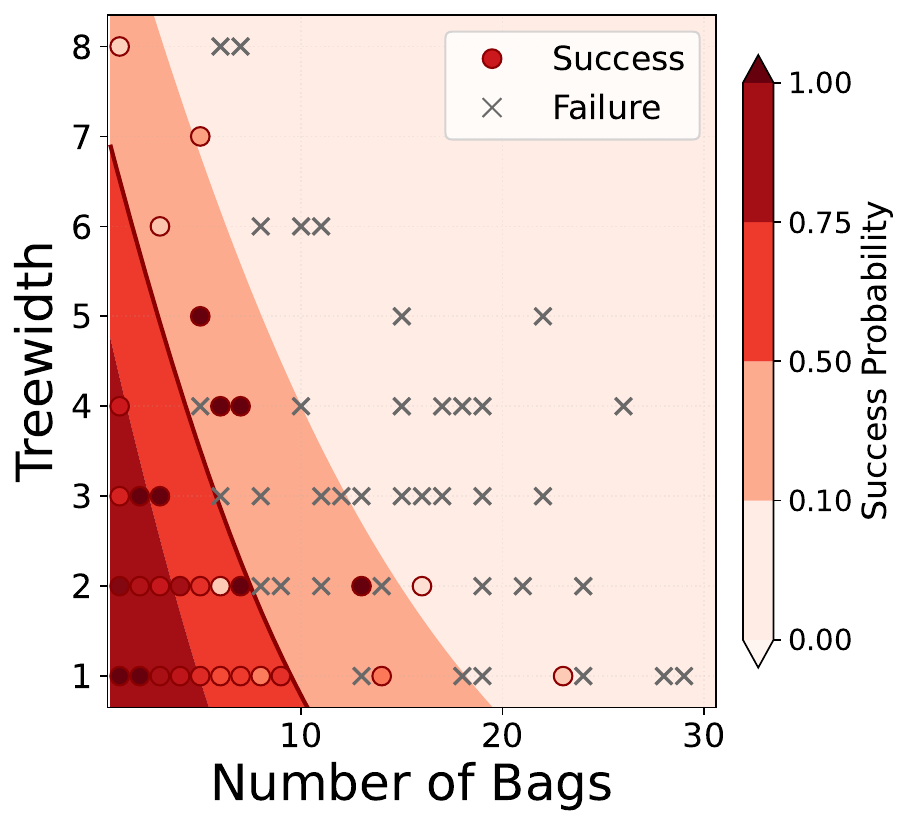}
        \caption{Claude baseline}
        \label{fig:claude_baseline}
    \end{subfigure}
    \hfill
    \begin{subfigure}[t]{0.49\linewidth}
        \centering
        \includegraphics[width=\linewidth]{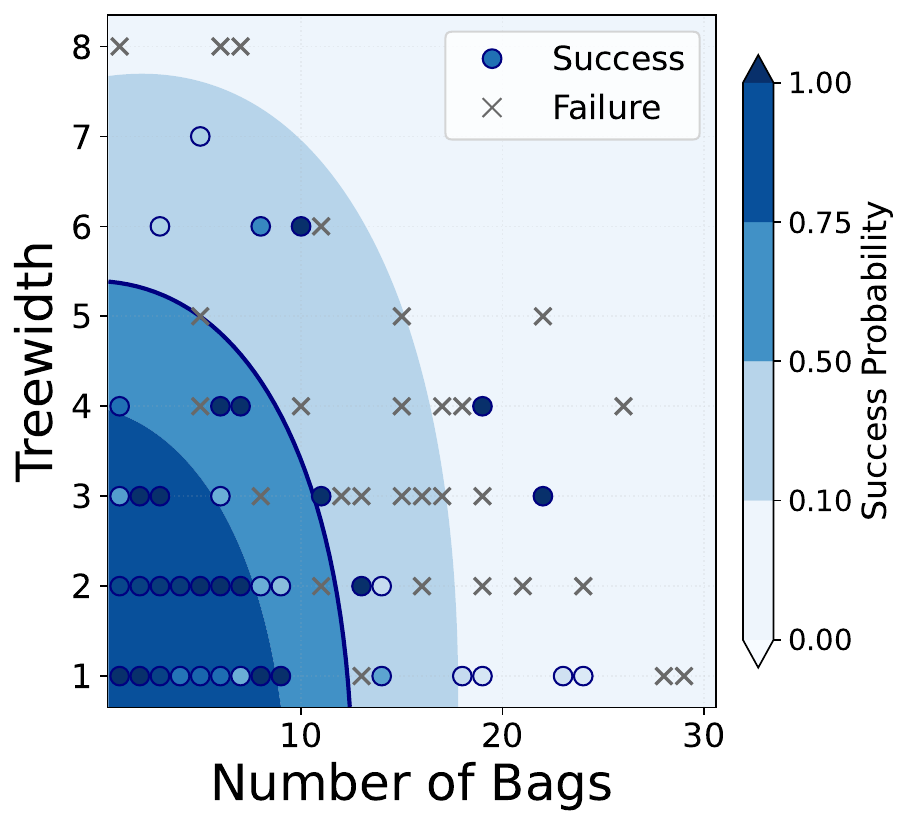}
        \caption{Claude \sys}
        \label{fig:claude_decomp}
    \end{subfigure}

    \vspace{0.5em}


    \centering
    \begin{subfigure}[t]{0.50\linewidth}
        \centering
        \includegraphics[width=\linewidth]{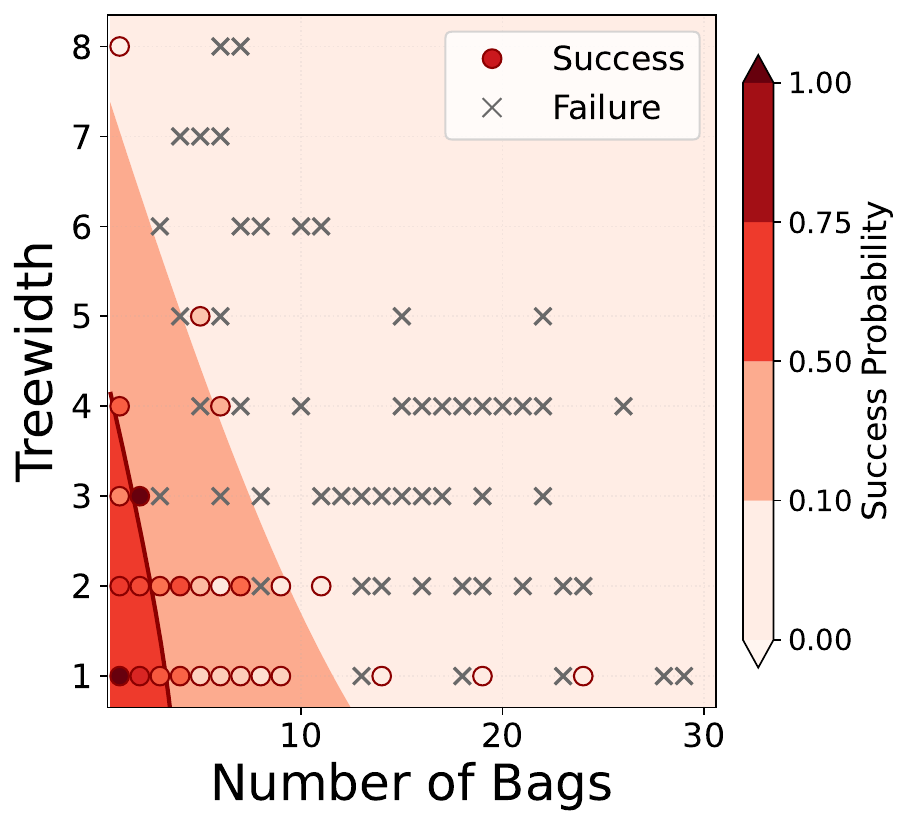}
        \caption{LLaMA baseline}
        \label{fig:llama_baseline}
    \end{subfigure}
    \hfill
    \begin{subfigure}[t]{0.48\linewidth}
        \centering
        \includegraphics[width=\linewidth]{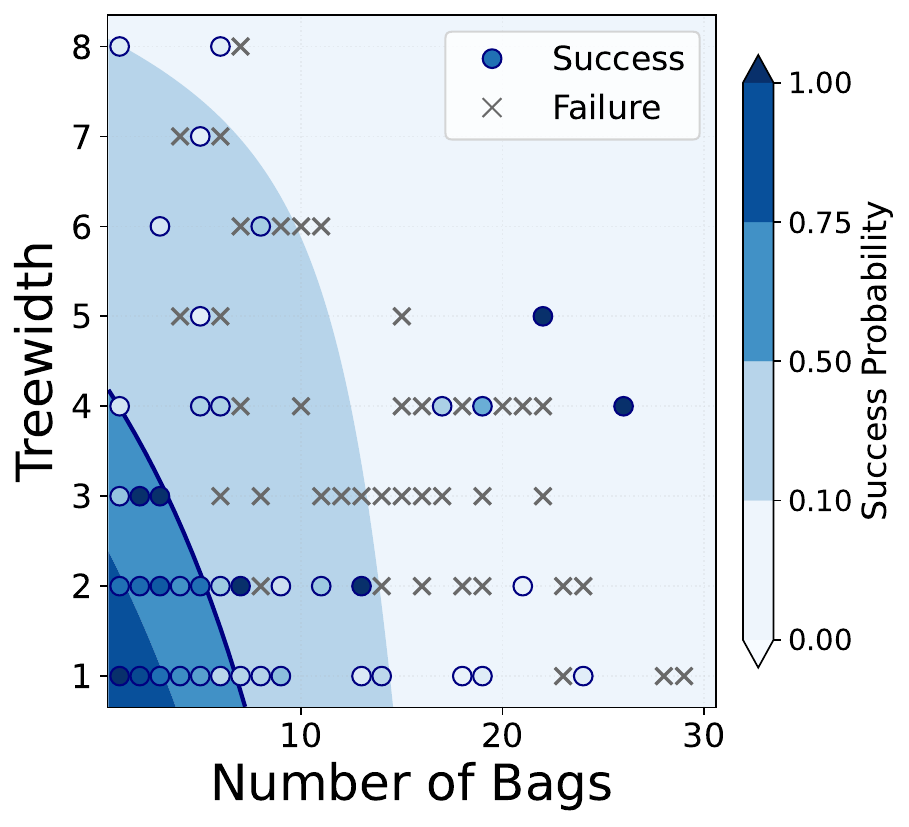}
        \caption{LLaMA \sys}
        \label{fig:llama_decomp}
    \end{subfigure}
    \caption{Our complexity measures define \textbf{frontiers of difficulty}.  SOTA models like Claude shifts the boundaries towards the right, while \sys's decomposition consistently pushes the frontiers towards more complex problems.  }
    \label{fig:llama_comparison}

\end{figure}




We evaluated Llama-3-70B on all tasks and Claude3.5-Sonnet on half the tasks (randomly sampled).   
\Cref{fig:llama_comparison} plots the success/failure of each task by its complexity measures, and exhibits frontiers of difficulty beyond which tasks are too difficult or too simple.   For the baseline, there appears to be a fixed ``total task complexity'' evidenced by the trade-off between problem treewidth and number of bags (left column).   In contrast, \sys decomposition pushes the frontier outward and is able to successfully complete more complex tasks.    Overall, \sys increases the task completion rate from $49.3\%\to 58.1\%$ using Claude, and $21.5\%\to 36.5\%$ using LLaMA---a $9-15\%$ improvement on both models.  


\begin{figure}
    \centering
    \includegraphics[width=.9\columnwidth]{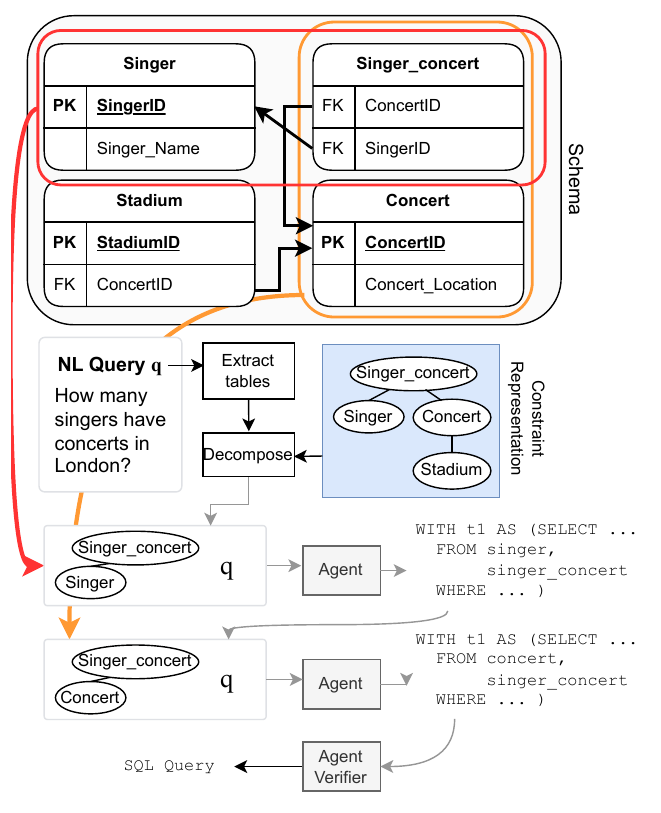}
    \caption{Instead of giving the full database schema and NL query $q$ to an agent, \sys decomposes the NL task into two subtasks, each given a subset of the database schema, $q$, and the output of the previous subtask if any.   }
    \label{fig:db_example}
\end{figure}

\subsection{Natural Language to SQL}

Natural language to SQL (NL2SQL) tasks translate a natural language query over a database into a desired SQL query.   A major challenge is constructing the join condition to connect tables referenced in the query.
For example, the schema in \Cref{fig:db_example} contains 4 tables about singers and concerts.   Although the NL query only refers to singers and concerts, the LLM needs to infer that  \texttt{singer\_in\_concert} is needed for a valid join path. 

The complexities of both the query (the number of tables referenced) and the database schema (number of tables and their foreign key graph) affect the agent's ability to construct queries with the appropriate join graphs.  In addition, database theory~\cite{gottlob2001hypertree} already models a database as a constraint graph (tables are nodes, and foreign keys are edges) and a query as a subgraph; this representation has also been used for NL2SQL approaches~\cite{wang2019rat}.   Thus, we use NL2SQL to study decomposition.


We use the popular Spider NL2SQL benchmark dataset~\cite{yu2019spiderlargescalehumanlabeleddataset}.  It contains hundreds of databases, each containing up to 37 tables, 90 foreign keys, and potentially over 100 columns.   Each task consists of a database schema $S$, the natural language (NL) query $q$, and a ground-truth SQL query $q^*$. 
The original benchmark submits $(S,q)$ to the LLM, which can result in incorrect or invalid join conditions.   Given the tables referenced in $q$, \sys decomposes the schema into subgraphs with minimized maximal complexities and constructs a workflow (\textcolor{grey}{grey arrows}).  For instance, the first subtask gives the agent $q$ and the schemas for subgraph \texttt{singer-singer\_concert} (\textcolor{red}{red arrow}), and asks it to construct the appropriate \texttt{WITH} clause.   The next subtask does the same with the second subgraph (\textcolor{orange}{orange arrow}), but is also provided the output from the previous subtask.   A verification agent cleans up minor errors to build the final SQL query.

\begin{figure}
    \centering
    \includegraphics[width=1\linewidth]{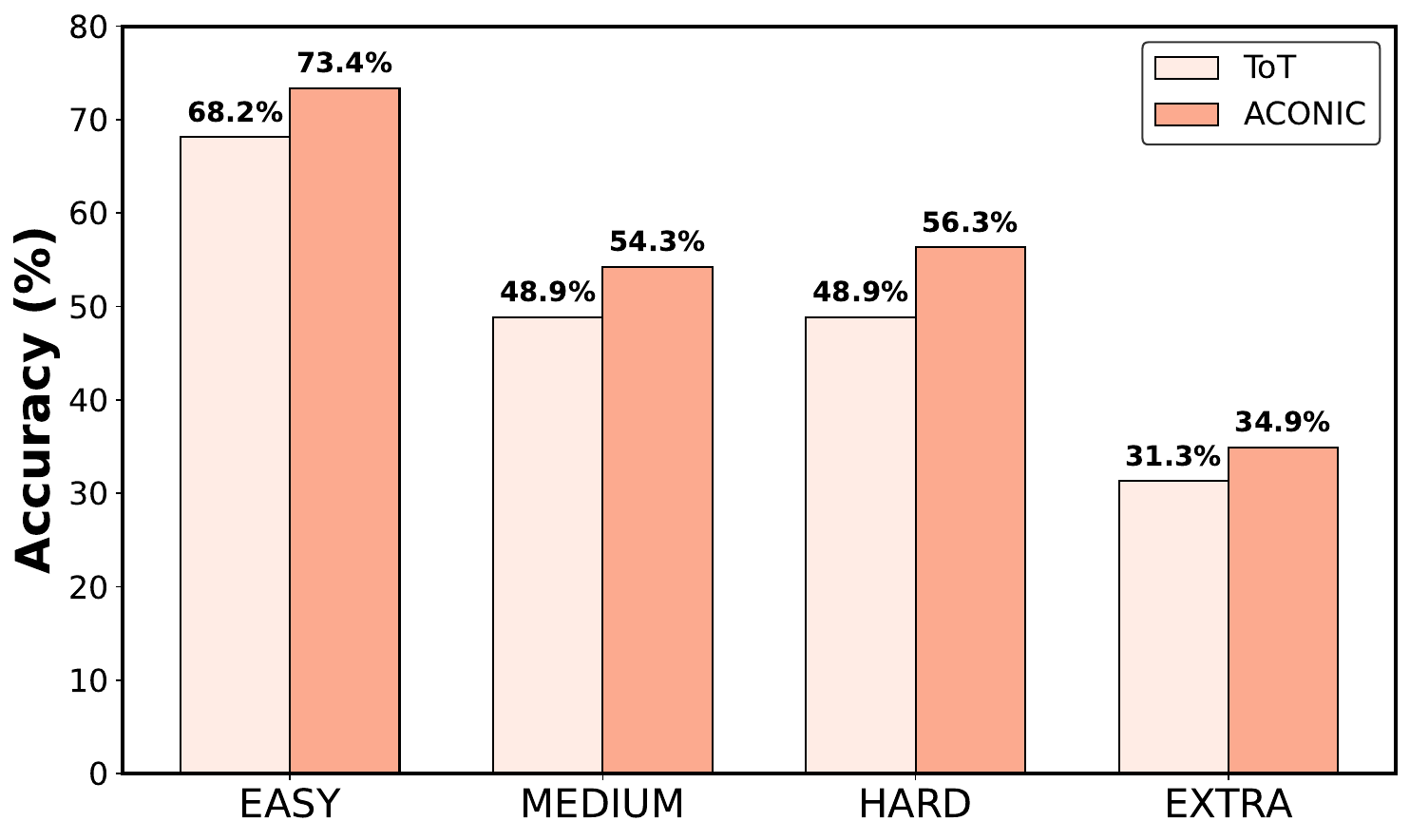}
    \caption{Accuracy performance of  Tree of Thought (ToT) and ACONIC (Ours)  on Spider EASY, MEDIUM, HARD, and EXTRA tasks}
    \label{fig:dbcot}
\end{figure}
\section{Conclusions}
\sys is a principled framework for decomposing complex LLM tasks by reducing them to constraint satisfaction problems and quantifying complexity via treewidth. Unlike heuristic methods, \sys minimizes local complexity while preserving global satisfiability. Across SATBench and Spider, this yielded \textbf{frontiers of task difficulty} that define LLM reasoning limits. Complexity-based decomposition improved completion rates by up to $15\%$ on SAT-bench. On Spider datasets, the accuracy was increased over ToT by up to 7.5 points (and +5.4 points overall).   These results suggest a path towards theoretically grounded, reliable multi-step LLM systems.
\Cref{fig:dbcot} reports Spider accuracy for \sys and Tree-of-Thoughts (ToT).
Across different difficulty levels, \sys consistently outperforms ToT by 3.6–7.5 accuracy points, with an overall gain of +5.4 points (50.68\% $\rightarrow$ 56.09\%).
\section{Limitations}
\sys provides a principled framework to analyze and reduce task complexity, but it does not yet constitute a fully autonomous decomposition or reasoning system.   It is also not intended to be a complete solution for general task decomposition, but rather as a theoretical and empirical examination of how \emph{constraint-induced complexity} affects the problem-solving ability of LLMs. 
 
Our focus has been on evaluating how complexity-guided decomposition impacts performance relative to heuristic baselines such as chain-of-thought prompting, rather than on direct comparison with other theoretically grounded or learning-based decomposition frameworks.  For this reason, we also do not claim that our final decomposition system is superior to existing solutions to the benchmark tasks.  For instance, SAT-bench can be directly solved using a constraint solver.  Similarly, the NL2SQL tasks can be readily solved using simple path-finding algorithms over the database schema.

The tasks that we evaluated can be conveniently modeled as constraint satisfiability problems.  Although many other practical problems, such as deadlock detection in databases, resource scheduling in operating systems, task placement in distributed systems, and control/data-flow construction in agentic programming, can be formulated as constraint satisfiability problems, they often cannot be  logically represented completely, either due to question ambiguity, lack of transparency of the agent actions, or fuzzy contextual information.   In these cases, future work might study hybrid decomposition approaches that mix logical and common-sense constraints in a task.   

\bibliography{custom}
\newpage
\appendix

\section{Appendix: SAT-Bench Transformation Details}

In this appendix, we provide a detailed explanation of how we perform the
formal reduction and experimental setup for the SAT-Bench benchmark. 
\subsection{Notation and Inputs}
A SAT-Bench instance provides:
\begin{itemize}[leftmargin=*, itemsep=0pt]
  \item A CNF over Boolean variables \(V=\{v_1,\dots,v_n\}\) with clauses \(\Phi=\{C_1,\dots,C_m\}\), where each clause with variable $\mathrm{vars}(C_j)=\{i_{j,1},\dots,i_{j,k_j}\}$ and $C_j=\bigl(\bigvee_{r=1}^{k_j}\ell_{j,r}\bigr)$ and each literal \(\ell_{j,r}\in\{v_i,\neg v_i\}\).
  \item A natural-language story \(Q\) and an alignment map \(\psi\) from story entities/conditions to SAT literals or short CNF fragments (as specified by SAT-Bench).
\end{itemize}
\subsection{CSP Encoding of SAT Clauses}
We build a CSP instance $\mathcal{R}=\langle X,D,C\rangle$ as follows.
\(X=\{x_1,\dots,x_n\}\) (one variable per SAT variable $v_i$) with domain of each variable being $D_{x_i}=\{0,1\}$

For each clause $C_j=\bigl(\bigvee_{r=1}^{k_j}\ell_{j,r}\bigr)$, let
$\mathrm{vars}(C_j)=\{i_{j,1},\dots,i_{j,k_j}\}$ be the indices of variables
appearing in $C_j$ (ignoring polarity). Define the \emph{literal evaluation}
map for an assignment $\mathbf{t}_j$ of all possible combinations in the domain $D$ ($\mathbf{t}_j\in D^{\mathrm{vars}(C_j)}$) to be with all local assignments to variables $vars(C_j)$ so that clause $C_j$ is satisfied. Therefore the constraint over the variables would be $\mathbf{T}_j$ with all valid $\mathbf{t}_j$. Each \(\mathbf{T}_j\) thus defines an $|\mathrm{vars}(C_j)|$-ary constraint relation over the variables appearing in \(C_j\). 

We encode the variable $V$ to CSP variables with initiation of 0 (False) and the clauses $C_j=\bigl(\bigvee_{r=1}^{k_j}\ell_{j,r}\bigr)$ with literal \(\ell_{j,r}\in\{v_r,\neg v_r\}\) as clause-wise constraint $\mathbf{T}_j$ including all the literals in $C_j$.

\subsection{Tree Decomposition on the Induced CSP Graph}
We represent each CSP variable as a node in the graph.  
For every constraint (clause) $C_j$ defined over the variable set $\mathrm{vars}(C_j)$,  
we connect all variables in $\mathrm{vars}(C_j)$ with edges, thereby forming a clique of size $|\mathrm{vars}(C_j)|$.  
The resulting undirected graph captures the variable–constraint dependencies of the SAT instance.

To analyze structural complexity, we perform \emph{tree decomposition} on this graph,  
which partitions the variable set into overlapping subgraphs (bags) while minimizing the size of the largest bag.  
The minimum achievable largest bag size minus one corresponds to the graph’s \emph{treewidth},  
which we use as a key measure of problem complexity and locality in constraint reasoning~\cite{bodlaender1998partial} in chart \ref{fig:claude_baseline}, \ref{fig:claude_decomp}, \ref{fig:llama_baseline}, \ref{fig:llama_decomp}

In our implementation, we compute the tree decomposition using  
\href{https://www.sagemath.org/}{\texttt{SageMath}},  
which provides efficient heuristics (e.g., minimum fill-in) and exact solvers for treewidth minimization.

\subsection{Experiment Workflow}

\begin{algorithm}[H]
\caption{SAT-Bench Workflow}
\begin{algorithmic}[1]
\STATE \textbf{Input:} NL story $Q$, CNF clauses $\Phi$, mapping $\psi$
\STATE \textbf{Build CSP:}
\STATE \hspace{1em} $X \leftarrow$ variables in $\Phi$
\FOR{each clause $C$ in $\Phi$}
    \STATE \hspace{1em} $T_C \leftarrow$ all satisfying assignments of variables in $C$
\ENDFOR

\STATE \textbf{Build constraint graph:}
\FOR{each clause $C$ in $\Phi$}
    \STATE \hspace{1em} connect variables in $C$ in graph
\ENDFOR

\STATE \textbf{Tree decomposition:}
\STATE \hspace{1em} Bags $\leftarrow$ tree\_decompose(graph)

\STATE \textbf{Agent interaction:}
\STATE \hspace{1em} State $\leftarrow$ empty assignment
\WHILE{not solved and budget not exceeded}
    \IF{CoT / ToT / DaC}
        \STATE \hspace{1em} Obs $\leftarrow$ (current violation, all clauses in $\Phi$)
    \ELSIF{ACONIC}
        \STATE \hspace{1em} Violated $\leftarrow$ clauses violated under State
        \STATE \hspace{1em} Bag $\leftarrow$ bag containing variables of Violated
        \STATE \hspace{1em} Obs $\leftarrow$ (current violation, clauses inside Bag)
    \ENDIF
    \STATE Action $\leftarrow$ LLM(State, $Q$, $\psi$, Obs)
    \STATE State $\leftarrow$ update assignment(Action)
\ENDWHILE

\STATE \textbf{Return} final assignment
\end{algorithmic}
\end{algorithm}
We detail here the experimental setup for both the baseline \textit{Chain-of-Thought} (CoT) decomposition and the proposed \sys{} method.  
The agent interacts with the environment over multiple rounds, progressively constructing a variable assignment.

At each round, the agent is provided with:
\begin{itemize}[leftmargin=*, itemsep=0pt]
  \item the natural-language query \(Q\),
  \item the variable-to-literal mapping \(\psi\),
  \item the current violation status (i.e., whether any clause in \(\Phi\) is unsatisfied under the current partial assignment), and
  \item a set of \textbf{observations}, which differ (and is the only difference) between the baseline and \sys{} configurations.
\end{itemize}

\paragraph{Baseline with full observation.}
In the Chain-of-Thought baseline, the agent receives the complete observation set,  
including all conditions (clauses) in the instance.  
This corresponds to a global reasoning setup in which the agent has full visibility of the constraint space.

\paragraph{\sys{} (Decomposed Observation).}
In the \sys{} configuration, the agent receives only the subset of conditions relevant to its current focus.  
Specifically, when a clause is violated, the system identifies the \emph{bag} in the tree decomposition that contains the variables of the violated clause,  
and provides the agent with all conditions (clauses) associated with that bag.  

Agents are required to provide a set of value assignment to the variables according to the observations, violation status, query, and the mapping $\psi$. The process ends until all the variables are properly set such that no violation is caused, or a maximum interaction budget is reached.

\subsection{Implementation}
We implement SAT-Bench experiment as a Terminal-Bench task~\cite{tbench_2025} to ensure reproducibility and comparability between agents.

\section{Appendix: Spider database decomposition details}

\subsection{Notation and Inputs}
The Spider benchmark~\cite{yu2019spiderlargescalehumanlabeleddataset} provides a large-scale collection of natural language (NL) queries paired with their corresponding database schemas, gold-standard SQL queries, and example data.  
In our experiments, we use only the NL query and the associated schema as inputs to the framework, the gold SQL and data are reserved for evaluation.

For queries involving multiple tables, we model the join operations as a \emph{constraint satisfaction problem} (CSP) defined over the Cartesian product of the participating tables.  
Each attribute join condition introduces a binary constraint between the respective table variables, and the feasible assignments correspond to tuples that jointly satisfy all join and filter predicates.

\subsection{CSP Encoding of Tables and Joins}
For the NL-to-SQL task, we model relational queries as constraint satisfaction problems (CSPs) defined over database tables.  
Each table in the schema is represented as a CSP variable, and its domain corresponds to the power set of all possible tuple combinations within that table:
\[
X_i = \text{Table}_i, \qquad D_i = 2^{\text{Tuples}(\text{Table}_i)}.
\]
Hence, a value assignment to \(X_i\) specifies a subset of rows (tuples) selected from that table under the query conditions.

For a join operation between two tables \(T_a\) and \(T_b\), the feasible results can be expressed as a subset of the Cartesian product of their tuple domains:
\[
D_{a,b} \subseteq D_a \times D_b,
\]
where each pair of tuples \((t_a, t_b)\) is retained if and only if it satisfies the join predicate (e.g., matching primary–foreign key attributes, SQL defined join rules).

In this formulation, the NL query is modeled as a set of \emph{semantic pairwise constraints} over tables and attributes.  
Each join, filter, or projection condition introduces a logical constraint between variable assignments.
Unary constraints are applied to represent filtering conditions on individual tables (e.g., \texttt{WHERE} clauses).  
Thus, solving the CSP corresponds to finding a consistent set of table subsets that jointly satisfy all query semantics implied by the NL input.

\subsection{Tree Decomposition on the Induced CSP Graph}
As the core structural analysis step in \sys{}, we perform tree decomposition on the CSP graph induced by the relational constraints.  
The procedure is analogous to that used for the SAT-Bench experiments.

We construct a constraint graph where each node corresponds to a table variable,  
and an undirected edge is added between two nodes if there is a Foreign Key constraint represented in the schema.

We then apply the tree decomposition algorithm that minimizes the maximum bag size to obtain a decomposition with minimal treewidth.  
The resulting tree structure defines a set of subproblems (bags), each representing a locally consistent subset of tables and constraints.  
As in the SAT-Bench setup, we use the \href{https://www.sagemath.org/}{\texttt{SageMath}} package to compute tree decompositions,  
leveraging its built-in heuristics (e.g., minimum fill-in) and exact solvers for treewidth minimization.
\subsection{Experiment Workflow}
We compare the baseline \textit{Tree-of-Thought} (ToT) heuristic decomposition method with the proposed \sys{} framework on the NL2SQL task.  
In both settings, the agent interacts with the environment over multiple rounds, constructing the final SQL query.
\begin{algorithm}[H]
\caption{Spider Workflow}
\begin{algorithmic}[1]
\STATE \textbf{Input:} NL query NLQ, database schema $S$

\STATE \textbf{Build CSP:}
\STATE \hspace{1em} $X \leftarrow$ tables in $S$

\STATE \textbf{Build constraint graph (foreign keys only):}
\FOR{each foreign key constraint}
    \STATE \hspace{1em} connect corresponding tables in graph
\ENDFOR

\STATE \textbf{Tree decomposition:}
\STATE \hspace{1em} Bags $\leftarrow$ tree\_decompose(graph)

\STATE \textbf{Agent interaction:}
\STATE \hspace{1em} CTEs $\leftarrow$ empty list
\WHILE{unresolved tables remain}
    \IF{ToT}
        \STATE \hspace{1em} Obs $\leftarrow$ (current SQL, full schema $S$)
    \ELSIF{ACONIC}
        \STATE \hspace{1em} Target $\leftarrow$ tables referenced in NLQ
        \STATE \hspace{1em} Bag $\leftarrow$ bag containing unresolved Target tables
        \STATE \hspace{1em} Obs $\leftarrow$ (CTEs, schema restricted to Bag)
    \ENDIF
    \STATE Fragment $\leftarrow$ LLM(NLQ, Obs, Schema $S$)
    \STATE append Fragment to CTEs
\ENDWHILE

\STATE \textbf{LLM merging:}
\STATE \hspace{1em} SQL $\leftarrow$ LLM\_based\_merge\_and\_resolve(CTEs)

\STATE \textbf{Return} final SQL
\end{algorithmic}
\end{algorithm}
At each round, the agent is provided with:
\begin{itemize}[leftmargin=*, itemsep=0pt]
  \item the natural-language query (NLQ),
  \item the database schema (whose visibility differs between the baseline and \sys{} setups), and
  \item the current intermediate assignment or generated SQL fragment.
\end{itemize}

\paragraph{Baseline (Full Schema).}
In the ToT baseline, the agent receives the entire database schema at once and is asked to generate the complete SQL query that satisfies the NLQ.  
This configuration represents global reasoning over the full relational structure, without any decomposition or locality constraints.

\paragraph{\sys{} (Decomposed Schema).}
In the \sys{} setup, the agent begins by identifying the \emph{target table(s)} relevant to the NLQ.  
This step allows \sys{} to eliminate irrelevant tables from the constraint graph, thereby reducing search space and improving reasoning focus.  
Given the current query context, the system retrieves the corresponding \emph{bag} in the tree decomposition that contains the undetermined tables.  
The agent is then provided with the local schema of this bag and is asked to generate a \emph{Common Table Expression (CTE)} that retrieves the subset of data relevant to the NLQ from these tables.

In subsequent rounds, the agent is prompted again with:
\begin{itemize}[leftmargin=*, itemsep=0pt]
  \item the original NLQ,
  \item the current decomposed schema (limited to the active bag and its boundary variables), and
  \item the CTEs generated from previous rounds.
\end{itemize}
This iterative process continues until all relevant tables have been resolved.  

Finally, a \emph{verification agent} aggregates and refines the partial results,  
merging the generated CTEs, resolving dependencies, and applying any remaining filters or ordering conditions to produce the final executable SQL statement.

\subsection{Implementation}
We implement Spider experiment as a Terminal-Bench~\cite{tbench_2025} task to ensure reproducibility and comparability between agents.

\section{Additional Analysis: Cost Evaluation}
\label{app:random_structure}

To assess more different methods and the token costs, we randomly selected tasks from SAT-Bench covering the joint distribution of treewidth and bag count induced by tree decomposition and tested with Chain of Thought (CoT), Divide and Conquer (DaC), Tree of Thoughts (ToT), and ACONIC(Ours). Specifically, we uniformly sample problem instances across discrete bins of $(\mathrm{tw}(G), |T|)$, where $\mathrm{tw}(G)$ denotes the treewidth of the constraint graph and $|T|$ the number of bags in the corresponding decomposition. This ensures that the evaluation set is not biased toward either very simple or very complex structural regimes.

Table~\ref{tab:random_structure_results} reports the aggregate accuracy and average token usage of each method under this sampling scheme.

\begin{table}[h]
\centering
\begin{tabular}{lcc}
\toprule
Method & Accuracy & Avg. Tokens \\
\midrule
CoT & 21.95\% & 24,179 \\
DaC & 21.95\% & 129,201 \\
ToT & 26.82\% & 209,976 \\
\textbf{ACONIC} & \textbf{29.27\%} & \textbf{19,247} \\
\bottomrule
\end{tabular}
\caption{Performance under random structural coverage across all treewidth and bag-count distributions.}
\label{tab:random_structure_results}
\end{table}

ACONIC achieves the highest accuracy while using fewer tokens than all baselines. This suggests that explicitly exploiting structural decompositions yields both effectiveness and efficiency benefits when reasoning over constraint-based tasks.

We also observe that Divide-and-Conquer (DaC) performs comparably to chain-of-thought in accuracy but incurs substantially higher token cost. This behavior is consistent with the nature of SAT and Spider tasks, which require maintaining and propagating multiple dependent variable assignments across steps. In such settings, naively decomposing the task into independent subquestions can lead to excessive context expansion without a corresponding gain in correctness.

Overall, these results support the view that structure-aware decomposition guided by the underlying constraint graph is better aligned with tasks characterized by strong variable dependencies, whereas generic divide-and-conquer strategies are less effective when subproblems are not weakly coupled.

\section{Additional Analysis: Tree of Thoughts}
\begin{figure}[t]

    \begin{subfigure}[t]{0.49\linewidth}
        \centering
        \includegraphics[width=\linewidth]{image/baseline_frontier_llama.pdf}
        \caption{Llama Chain of Thoughts}
        \label{fig:claude_baseline}
    \end{subfigure}
    \hfill
    \begin{subfigure}[t]{0.49\linewidth}
        \centering
        \includegraphics[width=\linewidth]{image/decomposition_frontier_llama.pdf}
        \caption{Llama Aconic \sys}
        \label{fig:claude_decomp}
    \end{subfigure}

    \vspace{0.5em}


    \centering
    \begin{subfigure}[t]{0.530\linewidth}
        \centering
        \includegraphics[width=\linewidth]{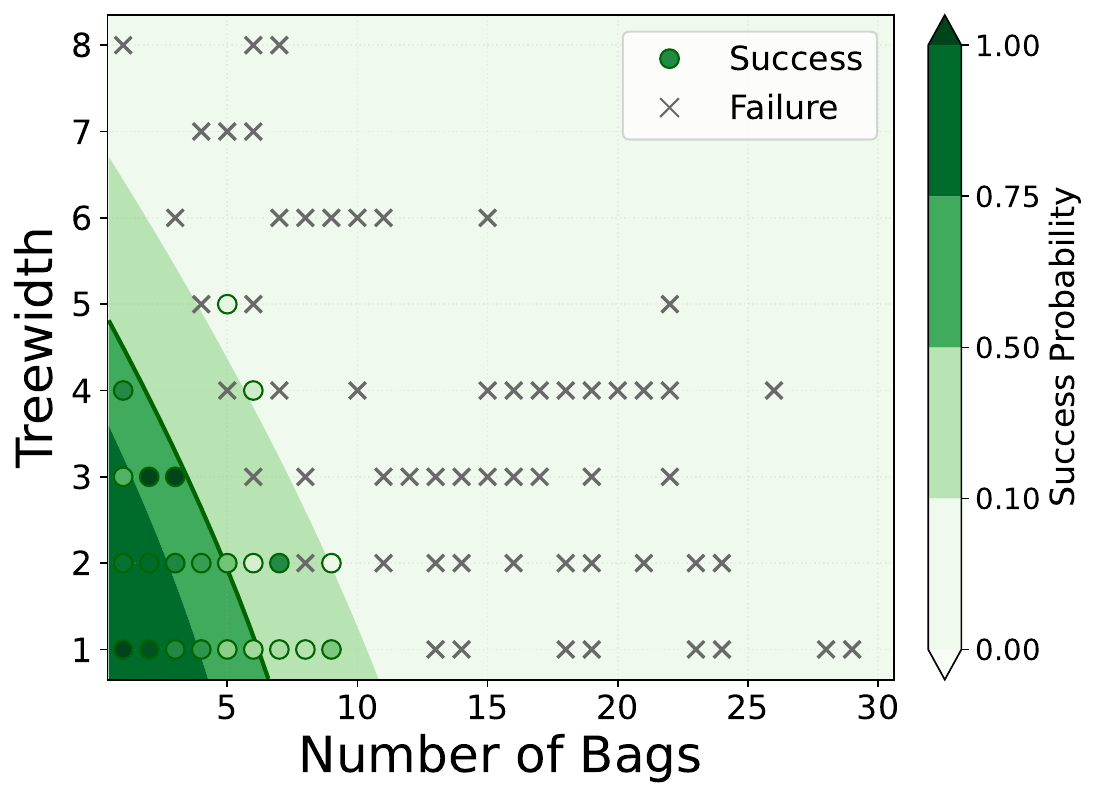}
        \caption{Llama Tree of Thoughts}
        \label{fig:llama_baseline}
    \end{subfigure}
    \hfill

    \caption{Task difficulty frontiers for LLaMA under CoT, ACONIC, and ToT on SAT-Bench.
Each subplot shows success rate distribution based on problem complexity.}
    \label{fig:llama_comparison_tot_added}

\end{figure}

We include results for Tree of Thoughts (ToT) on SAT-Bench (See figure \ref{fig:llama_comparison_tot_added}. Under our ToT configuration, ToT solves 34.8\% of tasks, compared to CoT's 21.5\% and ACONIC's 36.5\%.

\begin{table}[h]
\centering
\begin{tabular}{lccc}
\toprule
Method & CoT & ToT & ACONIC \\
\midrule
Success Rate & 21.5\% & 34.8\% & 36.5\% \\
\bottomrule
\end{tabular}
\caption{SAT-Bench success rate comparison across reasoning methods.}
\label{tab:tot_results}
\end{table}

We follow the implementation in the original paper~\cite{yao2023tree}. The search procedure is parameterized by: (i) maximum depth $T{=}5$, i.e., the search tree expands for at most 5 levels; (ii) branching factor $b{=}3$, i.e., each node proposes 3 candidate thoughts; and (iii) frontier size $k{=}5$, i.e., we keep the top 5 candidates at each depth as the frontier for further expansion.
\section{Boolean Encoding and SAT-to-CSP Reduction (Meeting Scheduling Example)}
\label{app:meeting_sat_to_csp}

We provide the mechanical steps that instantiate the meeting scheduling example as a Boolean formula and then as a CSP.

\paragraph{Variables and domains.}
Let $\mathcal{P}$ denote the set of participants, $\mathcal{T}$ the set of time slots, and $\mathcal{L}$ the set of locations.  
For each person $p\in\mathcal{P}$ and slot $(t,\ell)\in\mathcal{T}\times\mathcal{L}$, introduce a Boolean variable
\[
x_{p}^{(t,\ell)} \in \{0,1\},
\]
where $x_{p}^{(t,\ell)}=1$ indicates that $p$ attends slot $(t,\ell)$.  
For each person $p$, let $\mathcal{A}_p \subseteq \mathcal{T}\times\mathcal{L}$ be the feasible availability pairs extracted from the system record.  
For each desired meeting pair $(p,p')$, define the pairwise feasible set
\[
\mathcal{F}_{p,p'} := \mathcal{A}_p \cap \mathcal{A}_{p'}.
\]

\paragraph{Boolean formulation of constraints.}
Define meeting indicators for a pair $(p,p')$:
\[
M_{p,p'}^{(t,\ell)} := x_{p}^{(t,\ell)} \wedge x_{p'}^{(t,\ell)}.
\]
The task constraints consist of:
(i) \emph{Exactly-one feasible meeting slot} for each desired pair, and
(ii) No simultaneous meetings for a person. Concretely,
\begin{align}
&\bigwedge_{(p,p')\in\{(A,B),(A,C)\}} 
\mathrm{EO}\Big(\{ M_{p,p'}^{(t,\ell)} \mid (t,\ell)\in\mathcal{F}_{p,p'} \}\Big) \\
&\wedge \ \bigwedge_{(t,\ell)} 
\neg\Big(M_{A,B}^{(t,\ell)} \wedge M_{A,C}^{(t,\ell)}\Big),
\end{align}
where $\mathrm{EO}$ denotes \emph{Exactly One}.

\paragraph{SAT (CNF) representation.}
Let $\Phi=\{C_1,\dots,C_m\}$ denote the resulting CNF clauses, where each clause is a disjunction of literals over Boolean atoms (e.g., $x_{p}^{(t,\ell)}$ or $\neg x_{p}^{(t,\ell)}$). The meeting scheduling instance is satisfiable iff there exists an assignment to all involved Boolean atoms that satisfies every clause in $\Phi$. 

\paragraph{Reduction to CSP.}
Given $\Phi$, we construct a CSP $\mathcal{R}=\langle X,D,\mathcal{C}\rangle$ as follows.

\emph{Variables.} $X$ contains one CSP variable for each Boolean atom appearing in $\Phi$.
For notational simplicity, we reuse the same symbol for the CSP variable (e.g., $x_{p}^{(t,\ell)}$).

\emph{Domains.} Each variable is Boolean with domain
\[
D_x=\{0,1\}.
\]

\emph{Constraints.} For each CNF clause $C_j$ with variable set $\mathrm{vars}(C_j)\subseteq X$, we introduce $\mathbf{T}_j$ where $\mathbf{T}_j$ contains exactly those local assignments to $\mathrm{vars}(C_j)$ that satisfy clause $C_j$. Letting $\mathcal{C}=\{\mathbf{T}_1,\dots,\mathbf{T}_m\}$, an assignment satisfies the CSP iff it satisfies all clause relations.

\paragraph{Equivalence.}
By construction, any satisfying assignment of the CNF formula $\Phi$ induces a satisfying assignment of $\mathcal{C}$, and any satisfying assignment of $\mathcal{C}$ satisfies every clause in $\Phi$. Therefore, solving the resulting CSP is equivalent to solving the original meeting scheduling instance.

\end{document}